\documentclass[journal]{IEEEtran}

\ifCLASSINFOpdf
\else
   \usepackage[dvips]{graphicx}
\fi
\usepackage{url}

\usepackage{graphicx}

\usepackage{t1enc,amsmath,amsfonts,epsfig,graphicx,color,soul,mathabx,mathtools,listings,bm,amssymb}
\usepackage[latin1]{inputenc}
\usepackage{booktabs}
\usepackage[active]{srcltx} 
\newcommand{\eq}[1]{\begin{align}#1\end{align}}

\def\BibTeX{{\rm B\kern-.05em{\sc i\kern-.025em b}\kern-.08em
		T\kern-.1667em\lower.7ex\hbox{E}\kern-.125emX}}

\renewcommand{\baselinestretch}{1 } 

\everymath{\displaystyle}

\graphicspath{{./Figs/}{./fig/}}
\definecolor{violeta}{rgb}{0.2, 0, 0.8}
\definecolor{violeta-}{rgb}{0.5, .5, 1}

\newtheorem{definition}{Definition}
\newtheorem{proposition}{Proposition}

\newtheorem{theorem}{Theorem}

\input{def11.set}

\newcommand{\real}{\Re{e}}

\newcommand{\bi}[1]{\textbf{\textit{#1}}}

\begin{document}

\title{The alpha-beta divergence for real and complex data}

\author{Sergio Cruces, \IEEEmembership{Senior Member, IEEE}
\thanks{This work was supported in part by the MICIU/AEI/10.13039/5011 00011033 under Grant PID2021-123090NB-I00, and in part by ERDF/EU.}
\thanks{S. Cruces is with the Department of Signal Processing and Communications, Universidad de Sevilla, Spain (e-mail: sergio@us.es).}
}

\maketitle

\begin{abstract} 
Divergences are fundamental to the information criteria that underpin most signal processing algorithms. The alpha-beta family of divergences, designed for non-negative data, offers a versatile framework that parameterizes and continuously interpolates several separable divergences found in existing literature. This work extends the definition of alpha-beta divergences to accommodate complex data, specifically when the arguments of the divergence are complex vectors. This novel formulation is designed in such a way that, by setting the divergence hyperparameters to unity, it particularizes to the well-known Euclidean and Mahalanobis squared distances. Other choices of hyperparameters yield practical separable and non-separable extensions of several classical divergences. In the context of the problem of approximating a complex random vector, the centroid obtained by optimizing the alpha-beta mean distortion has a closed-form expression, which interpretation sheds light on the distinct roles of the divergence hyperparameters. These contributions may have wide potential applicability, as there are many signal processing domains in which the underlying data are inherently complex.
\end{abstract}


\IEEEpeerreviewmaketitle

\section{Introduction}
\IEEEPARstart{D}{ivergence} measures are fundamental building blocks for the design of loss functions and optimization criteria in signal processing, as they provide a mathematical framework to measure the dissimilitude between probability distributions~\cite{KL}. Generalized divergences extend beyond the classical limitation of considering probability distributions, allowing us to compare nonprobabilistic data representations~\cite{div-chapter}.

The alpha-beta divergences represent a generalization in the field of information-theoretic divergence measures with broad implications for machine learning and statistical signal processing \cite{BPAward}. In particular, they offer a flexible and powerful approach to analyzing non-negative datasets. Although initially proposed in~\cite{AB-div} within the context of nonnegative matrix factorization (NMF), they were extrapolated later in~\cite{AB-matrix} to allow comparison of positive definite matrices.
Since then, the applications of alpha-beta divergences have been transversal to several signal processing and machine learning fields, such as:  generative modelling~\cite{AB-GAN}, variational inference~\cite{AB-decomposable}-\cite{VariationalInference}, discriminative modelling \cite{AB-asymmetric-clustering}-\cite{AB-centroid}, dimensionality reduction~\cite{AB-SNE}, dictionary learning~\cite{AB-audio}, machine learning optimization~\cite{Access-2024}-\cite{AB-matrix-opt}, and metric learning on non-linear manifolds~\cite{AB-metric-learning1}-\cite{AB-metric-learning4}.

There are many signal processing applications that inherently rely on complex data representations, such as complex spectrograms in audio or equivalent low-pass signals and constellations in communications. Therefore, there is a real need for divergence functions that guide the criteria of shallow algorithmic models and deep neural networks in these domains \cite{C-DNN}-\cite{C-VAE}. Among the application of NMF to audio signal processing~\cite{NMF-review}, some works have bypassed this limitation with the proposal of a Complex Matrix Factorization framework \cite{C-NMF1}-\cite{C-NMF4}, which operates directly on complex-valued spectrograms. While these proposals were mostly limited to the Euclidean and Kullback-Leibler divergences, the work in \cite{complex-beta-div} proposes the ``Complex Beta divergence'', an interesting (although discontinuous) extension of the $\tilde{\beta}$-divergence  \cite{beta-div} that operates on complex values. As we will see, the alpha-beta divergence for complex values that we propose particularizes for $\alpha=1$ and $\beta\in \mathbf{R}$ to a different generalization of the $\tilde{\beta}$-divergence, where $\tilde{\beta}=1+\beta$, but which preserves the continuity with respect to $\tilde{\beta}$. Other choices of the pair of hyperparameters $(\alpha,\beta)\in\mathbf{R}^2$ provide an extension of several existing divergences for handling complex vectors. 

\section{notation}

This paper follows the standard notation where scalars are denoted by lowercase italic letters, and vectors are represented by lowercase bold letters. 
The generalized $(1-\alpha)$-exponential of $x\in\mathbf{R}$ (see \cite{Naudats} and Figure~1 of \cite{Cichocki2025MirrorDA}) is defined by 
\eq{ \label{defexp}
	\exp_{1-\alpha}(x)=\left\{
	\begin{array}{cl}
		\displaystyle [1+ \alpha\, x]_+^{1/\alpha} & \alpha\neq 0\\
		\exp(x) & \alpha=0
	\end{array}
	\right.
}
where $[\cdot]_+$ refers to the operator $\max\{0,\cdot\}$.

Divergences and squared distances between scalars or vectors will be denoted by the capital operator $D(\cdot,\cdot)$. In this sense, the Kullback-Leibler \cite{KL} or I-divergence \cite{I-div-unnormalized} between the non-negative scalars $|p|,|q|\in \mathbf{R}^+$ is represented by 
\eq{
	D_{KL}(|p|,|q|) = |p|\log \frac{|p|}{|q|} -|p|+|q| .
	\label{div-KL}
}
Similarly, the Itakura-Saito scalar divergence \cite{IS} is given by 
\eq{
	D_{IS}(|p|,|q|) = \left|\frac{p}{q}\right|-\log \left|\frac{p}{q}\right|-1 ,
	\label{div-IS}
}
and the squared Hellinger~\cite{Hellinger} distance (scaled by $4$) is 
\eq{
	D_{H}(|p|,|q|)
	= 2 \left(\sqrt{|p|}-\sqrt{|q|}\right)^2 \, .
	\label{div-H}
}

For real vectors $\bp,\bq\in\mathbf{R}^m_+$ of non-negative elements, separable versions of these divergences are also commonly considered in the existing literature
\eq{ 
	D_{sep}(\bp,\bq) 
	&= \sum_{i=1}^m D(p_i,q_i)  
	\label{div-sep}
}  
where, within the summation, $D(\cdot,\cdot) $ refers to the selected divergence for the comparison of scalar elements.

Although all the previous definitions only apply to non-negative values, the Euclidean distance applies to complex values. In this paper, we will try to fill this theoretical gap by proposing an alpha-beta divergence for complex arguments that continuously links several classical divergences with the Euclidean and Mahalanobis squared distances.   

Given a complex vector $\bp\in \mathbf{C}^m$, its length is represented by the norm operator $\|\bp\|$, and we denote the Hermitan-transpose vector by $\bp^H$. The normalized or principal vector is given by $\hat{\bp}=\bp/\|\bp\|$, while the Euclidean angle (see \cite{Angles in c}) between the complex vectors $\bp$ and $\bq$ is denoted by  
\eq{
	\angle_{pq}=\arccos\left(\real\left\{\hat{\bq}^H\hat{\bp}\right\}\right) \, .
	\label{angle_pq}
}

\section{Proposal of Alpha-Beta divergences for complex arguments}
In this section, we first introduce and briefly review the AB divergence proposed in \cite{AB-div} for non-negative and unnormalized measures. After that, we present a decomposition of the Euclidean divergence, which will later be key for understanding and interpreting the proposed non-separable and separable extensions of the AB divergence for complex vectors.

\subsection{The alpha-beta divergence for non-negative elements}

The AB divergence for non-negative scalars $|p|,|q|\in \mathbf{R}_+$ is given by 
\eq{ 
	D_{AB}^{(\alpha,\beta)}(|p|,|q|)
	&= \frac{
		-(\alpha+\beta)|p|^\alpha |q|^\beta
		+\alpha |p|^{(\alpha+\beta)}
		+ \beta |q|^{(\alpha+\beta)} 
	}{\alpha\beta(\alpha+\beta)}
	\label{div-AB}
}
provided its hyperparameters satisfy $\alpha,\beta,(\alpha+\beta)\neq 0$. As detailed in Appendix A of \cite{AB-div}, the fundamental inequality $D_{AB}^{(\alpha,\beta)}(|p|,|q|)\geq 0$ (being only equal to zero for $|p|=|q|$) stems from a synthesis three Young's inequalities, each applicable within a distinct region of the $(\alpha,\beta)$ plane. All singular cases where equation (\ref{div-AB}) is undefined (when $\alpha, \beta$ or $(\alpha+\beta)$ are zero) are addressed by the divergence's extension by continuity 
\eq{
	&D^{(\alpha,\beta)}_{AB} \left(|p| , |q|\right)
	\doteq \label{div-extension} \\
	&\hspace{.6cm} 
	\left\{
	\begin{tabular}{ll}
		$\displaystyle \frac{1}{\alpha^2} \left( |p|^\alpha \log \left|\frac{p}{q}\right|^\alpha - |p|^\alpha + |q|^\alpha \right)  $   
		&$\alpha \neq 0, \beta = 0$
		\\
		$\displaystyle 	\frac{1}{\beta^2}
		\left( |q|^\beta  \log \left|\frac{q}{p}\right|^\beta - |q|^{\beta} + |p|^{\beta}  \right)$
		& $\alpha = 0, \beta\neq 0$
		\\
		$\displaystyle \frac{1}{\alpha^2} 
		\left(\left|\frac{p}{q}\right|^\alpha - \log \left|\frac{p}{q}\right|^\alpha  -1 \right)$ 
		& $\alpha=-\beta \neq 0 $
		\\
		$\displaystyle \frac{1}{2} 
		\left(\log |p|- \log |q|\right)^2 $
		& $\alpha=\beta = 0 .$
	\end{tabular}
	\right. \nonumber
}
In order of appearance, the cases presented by (\ref{div-extension}) respectively correspond to the family of generalized KL divergences (when $\alpha \neq 0, \beta = 0$), the family of dual generalized KL divergences ($\alpha = 0, \beta\neq 0$), the family of generalized Itakura-Saito divergences ($\alpha=-\beta \neq 0$), and the log-Euclidean squared distance ($\alpha=\beta = 0$).

\begin{figure}
	\centerline{\includegraphics[width=\columnwidth]{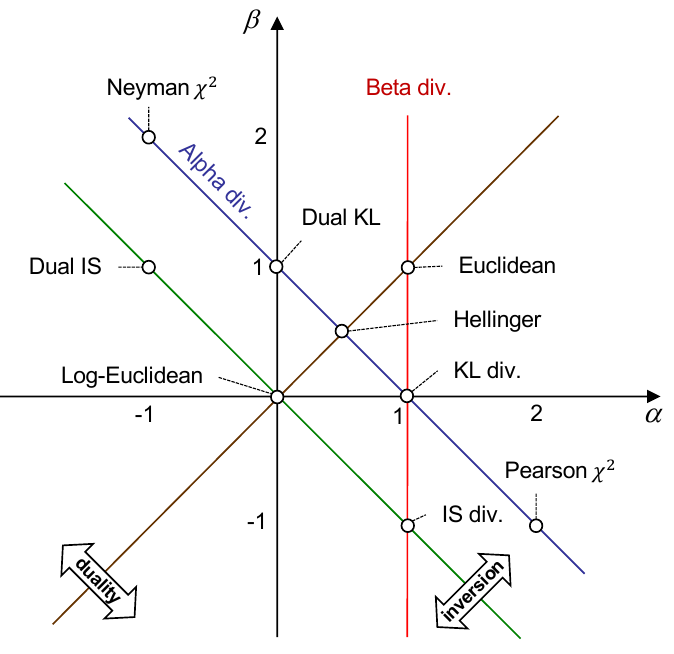}}
	\caption{Classical divergences for non-negative data as particularizations of the alpha-beta family of divergences together with their specific locations on the $(\alpha,\beta)$-plane. }
	\label{Fig1}
\end{figure}

Therefore, the three Young's inequalities summarized by~(\ref{div-AB}), along with their extension by continuity in (\ref{div-extension}), define an AB divergence between non-negative scalars across the entire $(\alpha,\beta)\in \mathbf{R}^2$ plane. In this sense, Figure~\ref{Fig1} illustrates classical divergences and squared distances at their specific locations on the plane, which is parameterized by the hyperparameters of the AB divergence. Specifically, the Euclidean divergence is obtained when $\alpha=\beta=1$, the KL divergence in (\ref{div-KL}) when $\alpha=1,\beta=0$, the IS divergence in (\ref{div-IS}) when $\alpha=-\beta=1$, the Hellinger divergence in (\ref{div-H}) is obtained when $\alpha=\beta=1/2$, and the log-Euclidean divergence when $\alpha=\beta=0$. For comparing not only scalars but also vector arguments with non-negative elements, separable versions of these divergences readily follow from~(\ref{div-sep}).

\subsection{Decomposition of the Euclidean divergence} 
Consider the definition of the Euclidean divergence between the complex vectors $\bp,\bq\in\mathbf{C}^m$, which is given by
\eq{ 
	D_{E}(\bp,\bq) 
	&=  \frac{1}{2} \|\bp-\bq\|^2. 
	\label{div-E-vector}
}
Given the representation of the complex vectors 
\eq{
	\bp=\|\bp\|\hat{\bp}\quad \text{and}\quad \bq=\|\bq\| \hat{\bq}
} 
in terms of their respective norms and direction vectors. We can additively decompose this divergence into the complementary additive contributions:
\eq{ 
	D_E(\bp,\bq) = D_{E}(\|\bp\|,\|\bq\|) + D_{\angle}(\bp,\bq),
	\label{E-decomp}
}
where the first term of this decomposition corresponds to the Euclidean divergence between norms, while the second term
\eq{ \label{div-E-angle}
	D_{\angle }(\bp,\bq) 
	&=  \|\bp\| \|\bq\|- \real\left\{\bq^H\bp\right\}\\
	&=  \|\bp\| \|\bq\|\left(1-\cos(\angle_{pq})\right)
}
accounts for the contribution resulting from the Euclidean angular discrepancy in (\ref{angle_pq}) between the direction vectors.
 
\subsection{Alpha-beta divergence for complex vectors}

There are two possible versions of the AB divergence for complex vectors. They depend on whether the comparison is performed directly on the lengths and principal directions of the involved vectors (which usually leads to a non-separable divergence) or, alternatively, if an accumulated and decoupled comparison is desired between the moduli and complex signs of the vector elements at the same positions (resulting in a separable divergence). In this subsection, we will start by introducing the former, "non-separable" case, as the separable version of this divergence will simply follow from~(\ref{div-sep}).

In similarity with the decomposition of the squared Euclidean distance in (\ref{E-decomp}) into two complementary contributions, we propose the following divergence extension:

\begin{definition}
For all $\bp,\bq\in \mathbf{C}^m$ we define the alpha-beta divergence as
\eq{ 
	D^{(\alpha,\beta)}(\bp,\bq) =  D_{AB}^{(\alpha,\beta)}(\|\bp\|,\|\bq\|) 
	+ D_{\angle}^{(\alpha,\beta)}(\bp,\bq) 
	\label{DibABcomplex}
}
where 
\eq{ 
	D_{\angle}^{(\alpha,\beta)}(\bp,\bq)
	&= \|\bp\|^{\alpha-1}\|\bq\|^{\beta-1} D_{\angle}(\bp,\bq)\\
	&= \|\bp\|^\alpha \|\bq\|^\beta (1-\cos(\angle_{pq} )) .
	\label{div08.}
}
\end{definition}
The first term in our definition (\ref{DibABcomplex}) is the divergence between vector norms,
while the second term, $D_{\angle}^{(\alpha,\beta)}(\bp,\bq)$, quantifies the divergence contribution stemming from the angular discrepancy between these vectors. This proposed divergence satisfies the fundamental inequality: 
\begin{proposition}
	$D^{(\alpha,\beta)}(\bp,\bq)$ is non-negative for all $\bp,\bq\in \mathbf{C}^m$, and $D^{(\alpha,\beta)}(\bp,\bq)=0$ if and only if $\bp=\bq$.
\end{proposition}	
The proof is trivial because both additive terms in the divergence definition are non-negative and are simultaneously zero only when the compared vectors coincide. 

 \begin{table}
 	\caption{Special cases of the Alpha-Beta divergence}
 	\label{Table1}
 	\small
 	\setlength{\tabcolsep}{1pt}
 	\begin{tabular}{cc}
 		\bf Divergence& 
 		$\bm{(\alpha,\beta)}$\ \bf div. particularization for $\bp,\bq\in \mathbf{C}^m$\\ 
 		\hline
 		&\\
 		Euclidean& 
 			$D^{(1,1)}(\bp,\bq) = \tfrac{1}{2} \|\bp-\bq\|^2$\\[4pt]
	  	KL div.& 
	  		$D^{(1,0)}(\bp,\bq) =\! 		D_{KL}(\|\bp\|,\|\bq\|)+\|\bp\|(1-\cos(\angle_{pq}))$\\[4pt]	 
	  	Dual KL & 
	  		$D^{(0,1)}(\bp,\bq) =\! D_{KL}(\|\bq\|,\|\bp\|)+\|\bq\|(1-\cos(\angle_{pq}))$\\[4pt]	
	  	Hellinger&
	  		$D^{(1/2,1/2)}(\bp,\bq) = 2\left(\|\bp\|^\frac{1}{2}-\|\bq\|^\frac{1}{2}\right)^2+\hspace{1.9cm}$\\[1pt]
			&\hspace{1.3cm} $\|\bp\|^\frac{1}{2}\|\bq\|^\frac{1}{2}(1-\cos(\angle_{pq}))$\\[4pt]	  		 	  		
	  	IS div. & 
	  		$D^{(1,-1)}(\bp,\bq) =\! D_{IS}(\|\bp\|,\|\bq\|)+\tfrac{\|\bp\|}{\|\bq\|}(1-\cos(\angle_{pq}))$\\[6pt]
	  	Dual IS & 
	  		$D^{(-1,1)}(\bp,\bq) =\! D_{IS}(\|\bq\|,\|\bp\|)+\tfrac{\|\bq\|}{\|\bp\|}(1-\cos(\angle_{pq}))$\\[4pt]	
	  	Pearson\,$\chi^2$ & 
		$D^{(2,-1)}(\bp,\bq) =\! \frac{(\|\bp\|-\|\bq\|)^2}{2\|\bq\|}+\!\tfrac{\|\bp\|^2}{\|\bq\|}(1-\cos(\angle_{pq}))\!$\\[8pt]	
		Neyman\,$\chi^2$ & 
		$\! D^{(-1,2)}(\bp,\bq) =\! \frac{(\|\bp\|-\|\bq\|)^2}{2\|\bp\|}+\!\tfrac{\|\bq\|^2}{\|\bp\|}(1-\cos(\angle_{pq}))\!$\\[8pt]		  
		Log-Euclid.& 
			$D^{(0,0)}(\bp,\bq) = \tfrac{1}{2} \left(\log\|\bp\|-\log\|\bq\|\right)^2+$\\[1pt]
		    &\quad\qquad $(1-\cos(\angle_{pq}))$\\[4pt]	
  		Alpha div. & 
  			$D^{(\alpha,1-\alpha)}(\bp,\bq) = D_{AB}^{(\alpha,1-\alpha)}(\|\bp\|,\|\bq\|)+$\ \ \\[1pt] 
  			&\hspace{3.1cm} $\|\bp\|^\alpha\|\bq\|^{1-\alpha}(1-\cos(\angle_{pq}))$\\[5pt]
   		Beta div. 
   		& 
   			$D^{(1,\tilde{\beta}-1)}(\bp,\bq) = D_{AB}^{(1,\tilde{\beta}-1)}(\|\bp\|,\|\bq\|)+$\\[1pt] 
			&\hspace{2.9cm} $\|\bp\|\|\bq\|^{\tilde{\beta}-1}(1-\cos(\angle_{pq}))$\\[5pt]		
  		AB div. & $D^{(\alpha,\beta)}(\bp,\bq) = D_{AB}^{(\alpha,\beta)}(\|\bp\|,\|\bq\|)+$\ \ \\[1pt] 
			&\hspace{2.9cm} $\|\bp\|^\alpha\|\bq\|^\beta(1-\cos(\angle_{pq}))$\\[6pt]			  		 
  		Separable & 
  			$D_{\text{sep}}^{(\alpha,\beta)}(\bp,\bq)  = \sum_{i=1}^m D^{(\alpha,\beta)}(p_i,q_i)$\\[5pt]	
  		Dual div. & \hspace{-.7cm}
  			$D^{(\beta,\alpha)}(\bp,\bq)
  			= D^{(\alpha,\beta)}(\bq,\bp)$\\[5pt]	
  		Inversion & 
  		$D^{(-\alpha,-\beta)}(\bp,\bq)
  		=
  		D^{(\alpha,\beta)}(\|\bp\|^{-1}\hat{\bp},\|\bq\|^{-1}\hat{\bq})$\\[5pt]		
  		Weighting & 
  			$D^{(\alpha,\beta)}_{\bf W}(\bp,\bq) =  D^{(\alpha,\beta)}(\bW^\frac{1}{2}\bp,\bW^\frac{1}{2}\bq)$\\[5pt]	
 		Mahalanobis& \ \ \   
 			$D^{(1,1)}_{\bm{\Sigma}^{-1}}(\bp,\bq) = \tfrac{1}{2} (\bp-\bq)^H \bSigma^{-1} (\bp-\bq)$ \\
 		&\\	
 		\hline	
		\\
 		\multicolumn{2}{c}{
 		where 
 		$\ \cos(\angle_{pq})\equiv\real\left\{\tfrac{\bq^H\bp}{\|\bq\|\|\bp\|}\right\}$}\\
 		\\
 		\hline
 	\end{tabular}
 	\label{tab1}
 \end{table}

A comprehensive summary of the special cases and properties of the  alpha-beta divergence for comparing complex vectors is presented in Table~\ref{Table1}. Depending on the case, the left column of the table refers to either the considered property or the classical divergence used to compare the norm of the involved vectors, while the right column shows either the corresponding properties or the particularizations of the alpha-beta divergence for complex vector comparison. 

The table presents the extension of the duality, inversion and weighting properties of the divergence to the complex case. The dual divergence $D^{(\beta,\alpha)}(\bp,\bq)$, which reverses the order of the hyperparameters, equals $D^{(\alpha,\beta)}(\bq,\bp)$, the original divergence with the arguments reversed. The divergence with opposite hyperparameters $D^{(-\alpha,-\beta)}(\bp,\bq)$ corresponds to the original divergence $D^{(\alpha,\beta)}(\|\bp\|^{-1}\hat{\bp},\|\bq\|^{-1}\hat{\bq})$ but applied to the vectors with a norm inversion.
Additionaly, for any $\bW$ that belongs to ${\cal P}_m$ (the open convex cone of Hermitian positive definite matrices), we define the weighted divergence $D_{\bf W}^{(\alpha,\beta)}(\bp,\bq)$ as $D^{(\alpha,\beta)}(\bW^\frac{1}{2}\bp,\bW^\frac{1}{2}\bq)$, which directly leads to the following result.
\begin{proposition}	
When  $\alpha=\beta=1$, the alpha-beta divergence for the comparison of $\bp,\bq\in \mathbf{C}^m$ guarantees consistence with both the Euclidean and Mahalanobis squared distances, since $D^{(1,1)}(\bp,\bq) = D_{E}(\bp,\bq)$ and, for $\bm{\Sigma}^{-1}\in {\cal P}_m$, we have 
\eq{
	D_{\bm{\Sigma}^{-1}}^{(1,1)}(\bp,\bq) = \tfrac{1}{2} (\bp-\bq)^H \bSigma^{-1} (\bp-\bq). 
}
\end{proposition}

\subsection{Alpha-beta divergence means and theirs centroids}
Consider the random vector $\bx$, which takes values in the discrete sample space ${\cal X}=\{\bx_1,\ldots,\bx_N\}$, where $\bx_n\in \mathbf{C}^m$. Its probability space is $({\cal X},{\cal F}, \nu)$, where ${\cal F}$ refers to the $\sigma$-algebra or event space, and $\nu$ is the probability measure.
Consider also the $\alpha$-transformed representation of $\bx$
\eq{
	\bx^{(\alpha)} =\|\bx\|^{\alpha-1}\bx ,
}
and the following statistics:
\eq{
	E[\bx^{(\alpha)}]&=  \sum_{n=1}^N \left(\nu_n \|\bx_n\|^{\alpha-1}\right)\bx_n, \\
	\left(E\left[\|\bx^{(\alpha)}\|\right]\right)^\frac{1}{\alpha}&=  \left(\sum_{n=1}^N \left(\nu_n \|\bx_n\|^{\alpha-1}\right)\|\bx_n\| \right)^\frac{1}{\alpha}.
	\label{Gmean0}
}
Here, the contributions of $\bx_n$ and of $\|\bx_n\|$ have been emphasized by the factors $\nu_n \|\bx_n\|^{\alpha-1}$.
Note that (\ref{Gmean0}) can be recognized as the weighted generalized $\alpha$-mean:
\eq{
	M_\alpha(\|\bx\|) 
	&=
	\left\{
	\begin{array}{cl}
		\displaystyle \left( \sum_{n=1}^N \nu_n \|\bx_n\|^\alpha  \right)^{\frac{1}{\alpha}}  & \alpha\neq 0\\
		\prod_{n=1}^N \|\bx_n\|^{\nu_n}  & \alpha=0
	\end{array}
	\right.
}
which has been extended by continuity for the case of $\alpha=0$.

\begin{theorem}
	The problem of minimizing the expected alpha-beta distortion when approximating the realizations of $\bx$ by a fixed and right-sided representative vector $\bc$, i.e.,
	\eq{
	\min_{\bi{c}\in \mathbf{C}^m} E[D^{(\alpha,\beta)}(\bx,\bc) ]  
	\equiv 
	\min_{\bi{c}\in \mathbf{C}^m} \sum_{n=1}^N \nu_n D^{(\alpha,\beta)}(\bx_n,\bc)
	}
	has the unique centroid solution 
	\eq{
	\bc_\star^{(\alpha,\beta)} &= \displaystyle 
		M_\alpha(\|\bx\|)\, 
		\exp_{1-\alpha}\left(- \beta\,     \xi_\alpha(\bx)\right)\frac{E[\bx^{(\alpha)}] }{\|E[\bx^{(\alpha)}]\|}\, ,
	\label{centroid-vector}
	}
	where $M_\alpha(\|\bx\|)$ is the generalized $\alpha$-mean of $\|\bx\|$ and
	\eq{
	\xi_\alpha(\bx)
	=\frac{E\left[\|\bx^{(\alpha)}\|\right]-\left\|E[\bx^{(\alpha)}]\right\|}{E\left[\|\bx^{(\alpha)}\|\right]}
	\in [0,1]
	}
	denotes the normalized Jensen's gap for the 2-norm, which is a dimensionless variable that quantifies the signed $\alpha$-directional deviation (or lack of concentration) of $\bx$ from a single ray from the origin. By duality, the minimization of the reverse expected divergence $E[D^{(\alpha,\beta)}(\bc,\bx) ]$ leads to the left-sided centroid solution $\bc_\star^{(\beta,\alpha)}$. 
\end{theorem}
The proof of this theorem is provided in the Appendix. 

When noise and outliers perturb the norm and alignment of the vector samples, the degrees of freedom provided by the divergence's hyperparameters can be exploited to achieve a certain degree of robustness. This is a consequence of the different roles played by $\alpha$ and $\beta$ in the resulting centroid solution. 

On the one hand, from (\ref{centroid-vector}), it is apparent that the normalized vector $\hat{\bc}_\star^{(\alpha,\beta)}=\bc_\star^{(\alpha,\beta)}/\|\bc_\star^{(\alpha,\beta)}\|$ only depends on $\alpha$, which controls the influence of the sample $\bx_n$ in the resulting principal centroid direction (through the weighting factors $\nu_n \|\bx_n\|^{\alpha-1}$). Samples with larger vector norms have a stronger influence for $\alpha>1$ and a weaker influence for $\alpha<1$, while the situation is the opposite for smaller vector norms.
 
On the other hand, the centroid's norm,  $\|{\bc}_\star^{(\alpha,\beta)}\|= M_\alpha(\|\bx\|)\, 
\exp_{1-\alpha}\left(- \beta\, \xi_\alpha(\bx)\right)$, depends on both $\alpha$ and $\beta$. Here, $\alpha$ controls the generalized mean, with positive values of $\alpha$ biasing $M_\alpha(\|\bx\|)$ towards the maximum of $\|\bx\|$ (obtained for $\alpha=\infty$), and negative values biasing it towards the minimum (obtained for $\alpha=-\infty$). Additionally, the term $\exp_{1-\alpha}\left(- [\beta\, \xi_\alpha(\bx)]\right)$ determines how the lack of sample alignment, quantified through $\xi_\alpha(\bx)$, influences the centroid norm. While $\exp_{1-\alpha}\left(-[\cdot]\right)$ is a monotonically decreasing function, whose convexity (or concavity) increases for $\alpha<1$ (or $\alpha>1$), the sign and scale of its argument are determined by $\beta$. The samples' misalignment  decreases the centroid norm progresively with $|\beta|$ for $\beta>0$, and increases it progresively for $\beta<0$, while for $\beta=0$, the norm simplifies to $M_\alpha(\|\bx\|)$, becoming independent of sample misalignment. 
  
\subsection{Separable version of the alpha-beta divergence}
The separable version of the divergence for the cumulative element-wise comparison of complex vectors $\bp$ and $\bq$ simply follows from~(\ref{div-sep}), resulting in the expression
\eq{ \label{separable}
	D_{sep}^{(\alpha,\beta)}(\bp,\bq)
	\ =&\ \ \sum_{i=1}^m D^{(\alpha,\beta)}(|p_i|, |q_i|) \nonumber\\
	 &+ \sum_{i=1}^m |p_i|^\alpha |q_i|^\beta  (1-\cos(\angle_{p_iq_i})
}
where $\angle_{p_iq_i}=\theta_{p_i}-\theta_{q_i}$, and $\theta_{(\cdot)}$ denotes the phase of each element. Similar to Table~1, the expression can be particularized for several values of $(\alpha,\beta)\in \mathbf{R}^2$. The first summation on the right-hand side of (\ref{separable}) compares the modulo elements and coincides with several classical separable divergences for non-negative measures for specific pairs of  $(\alpha,\beta)$, such as KL, IS, Hellinger, log-Euclidean, Neyman $\chi^2$ and Pearson $\chi^2$ divergences. The last summation, conversely, extends the divergence to deal with the phases of the compared elements.

A weighted separable divergence also follows from the statistical setting of the previous section when considering the case of a one-dimensional complex random variable
$x\in\mathbf{C}$. The expected alpha-beta distortion when approximating the realizations of $x$ by a fixed representative scalar $c\in\mathbf{C}$ adopts the form of the separable divergence
\eq{
	E[D^{(\alpha,\beta)}(x,c) ]  
	\equiv 
	\sum_{n=1}^N \nu_n D^{(\alpha,\beta)}(x_n,c)
}
whose optimization leads to the following right-sided centroid: 
\eq{
	c_\star^{(\alpha,\beta)} &= \displaystyle 
		M_\alpha(|x|)\, 
		\exp_{1-\alpha}\left(- \beta\, \xi_\alpha(x)\right)\frac{E[x^{(\alpha)}]}{|E[x^{(\alpha)}]|}\, . 
	\label{centroid-scalar}
}

When the random variable $x$ is non-negative, the elements of its sample space $x_n\in\mathbf{R}_+$ are trivially aligned, thus the relative alignment error $\xi_\alpha(x)$ is zero and $E[x^{(\alpha)}]=|E[x^{(\alpha)}]|$. This fact further simplifies the centroid's expression, making it independent of $\beta$ and reducing it to the generalized $\alpha$-mean of the non-negative random variable 
\eq{
	c_\star^{(\alpha,\beta)} &= \displaystyle 
	M_\alpha(x) 
	\,  .
\label{centroid-scalar-non-negative}
}
This result aligns with known centroids in the literature for several classical divergences. For instance, for the Euclidean, separable KL, and IS divergences, where $\alpha=1$, the centroid solution is the arithmetic mean $M_1(x)$. For the separable log-Euclidean and dual KL divergences, with $\alpha=0$, the centroid is the geometric mean $M_0(x)$. Lastly, for the separable dual IS divergence, where $\alpha=-1$, the centroid is the harmonic mean $M_{-1}(x)$.

An extended version of this manuscript will present our ongoing work on the natural extension of the alpha-beta divergence for comparing general complex matrices $\bP,\bQ\in \mathbb{C}^{m\times n}$.
 
%
 \section*{Acknowledgment}
 The author would like to thank the past collaborations and interesting discussions on this topic with professors A.~Cichocki, A.~Sarmiento, and S.i~Amari.
 %


 
\pagebreak

\newpage
\onecolumn
\appendix
\section*{Determining the right-sided and left-sided centroids for the expected alpha-beta divergence.}
 
The problem of minimizing the expected divergence 
\eq{
	{\cal D}(\bi{c}) = E[D^{(\alpha,\beta)}(\bx,\bi{c}) ] 
	\label{EDist}
}
arises when we approximate the realizations of a random vector $\bx$ with a single, right-sided representative vector $\bi{c}\in \mathbf{C}^m$. For this optimization, we express  $\bi{c}$ in its magnitude-direction representation, where $\bi{c}=\|\bi{c}\|\hat{\bi{c}}$.  
To minimize this divergence with respect to the unit norm vector $\hat{\bi{c}}$, we define the Lagrangian function
\eq{
	L(\hat{\bc},\lambda)= E[D^{(\alpha,\beta)}(\bx,\|\bc\|\hat{\bi{c}}) ] +\lambda(\|\hat{\bc}\|^2-1)\, .
}
Setting the gradient of the Lagrangian with respect to the complex conjugate of the unit-norm vector $\hat{\bi{c}}$ to zero,
\eq{
	\nabla_{\hat{\bi c}^*} L 
	= E[\nabla_{\hat{\bi c}^*} D^{(\alpha,\beta)}_\angle(\bx,\|\bc\|\hat{\bi{c}}) ] + \lambda\, \hat{\bi{c}} ={\bf 0} 
}
yields a parameterized family of candidate solutions 
\eq{
	\hat{\bi{c}}(\lambda) = -\frac{E[\nabla_{\hat{\bi c}^*} D^{(\alpha,\beta)}_\angle(\bx,\|\bc\|\hat{\bi{c}}) ]}{\lambda}
	= \frac{\|\bc\|^{\beta}}{\lambda} E[ \bx^{(\alpha)}]
	\label{family}
}
where $\bx^{(\alpha)}=\|\bx\|^{\alpha-1}\bx$.

The unit-norm constraint on $\hat{\bi{c}}(\lambda)$ allows us to determine the magnitude of the Lagrange multiplier, $|\lambda_\star|= \|\bc\|^{\beta}\|E[ \bx^{(\alpha)}]\|$. Substituting this magnitude into (\ref{family}) yields two critical points, $\hat{\bi{c}}(|\lambda_\star|)$ and $\hat{\bi{c}}(-|\lambda_\star|)$. Using the Cauchy-Schwarz inequality, it can be shown that these points correspond to the minimum and maximum, respectively, of the constrained distortion function. The minimum, which aligns with the signed-direction of the centroid solution, is therefore given by
\eq{
 	\hat{\bi{c}}_\star =\hat{\bi{c}}(|\lambda_\star|)= \frac{E[\bx^{(\alpha)}] }{\|E[\bx^{(\alpha)}]\|}\, .
} 

When this optimal direction $\hat{\bi{c}}_\star$ is substituted into (\ref{EDist}), the subsequent optimization of $E[D^{(\alpha,\beta)}(\bx,\|\bc\|\hat{\bi{c}}_\star) ]$ with respect to $\|\bc\|$ (subject to $\|\bc\|\geq0$) yields the following optimal centroid's length
\eq{
	\|\bc_\star\| &= \displaystyle 
	M_\alpha(\|\bx\|)\, 
	\exp_{1-\alpha}\left(- \beta\,\xi_\alpha(\bx)\right)\, .
	\label{norm_star}
}
To verify this, we first note that the feasible domain for $\|\bc\|$ is the interval ${\cal F}=[0,\infty)$. We then consider the case where the constraint $\|\bc\|\geq 0$ is inactive,  meaning the optimal solution lies in the interior of the domain. In this case, it can be shown that the expected divergence is equal to the alpha-beta divergence of the optimal length $\|\bc_\star\|$ from a candidate length $\|\bc\|$, plus a non-negative constant term
\eq{
	E[D^{(\alpha,\beta)}(\bx,\|\bc\|\hat{\bi{c}}_\star) ] 
	= D^{(\alpha,\beta)}(\|\bi{c}_\star\|, \|\bc\|)+ \text{const}
} 
where the constant term is equal to
\eq{
	\text{const} =
	\begin{array}{ll}
	\frac{1}{\beta}\left(\log_{1-(\alpha+\beta)} {\cal M}_\alpha(\|\bx\|)-\log_{1-(\alpha+\beta)}\|\bc_\star\|\right)\geq 0 & \text{for }\beta \neq 0.
	\end{array}    
} 
By continuity, the limit of this non-negative constant for $\beta=0$ must also be non-negative. Therefore, as long as $\|\bc_\star\|>0$, the result in (\ref{norm_star}) is the unconstrained minimum of the function.
Conversely, when the constraint is active, the solution must lie on the boundary $\partial {\cal F}$ of the feasible domain. In our case, the only element of the boundary is $\|\bc_\star\|=0$. Therefore, both the constrained and unconstrained optimizations yield the optimal centroid length given in (\ref{norm_star}).

Thus, the minimization of $\mathcal{D}(\bi{c})$ with respect to $\bc\in\mathbf{C}^m$ produces the product $\|\bc_\star\|\hat{\bi{c}}_\star$, which defines the right-sided centroid solution as
\eq{
 	\bc_\star^{(\alpha,\beta)} 
 	&= \displaystyle 
 	M_\alpha(\|\bx\|)\, 
 	\exp_{1-\alpha}\left(- \beta\,     \xi_\alpha(\bx)\right)\frac{E[\bx^{(\alpha)}] }{\|E[\bx^{(\alpha)}]\|}\, .
 	\label{centroid-vector-r}
}
Due to the duality property, the expected divergence remains invariant when both the arguments and hyperparameters are reversed, meaning $E[D^{(\alpha,\beta)}(\bc,\bx) ] = E[D^{(\beta,\alpha)}(\bx,\bc) ]$. Consequently, the left-sided centroid solution that minimizes the alpha-beta divergence is given by the expression for the right-sided centroid, but with the hyperparameters swapped. This yields the following solution for the left-sided centroid 
\eq{
	\bc_\star^{(\beta,\alpha)} &= \displaystyle 
	M_\beta(\|\bx\|)\, 
	\exp_{1-\beta}\left(- \alpha\,     \xi_\beta(\bx)\right)\frac{E[\bx^{(\beta)}] }{\|E[\bx^{(\beta)}]\|}\, .
	\label{centroid-vector-l}
}

\end{document}